# Canoe Paddling Quality Assessment Using Smart Devices: Preliminary Machine Learning Study

Shubham Parab[1], Anthony Lamelas[2], Ahmed Hassan[2], Pasang Bhote[2]

1. University of Michigan, Ann Arbor, MI    2. New York University, Brooklyn, NY

## Abstract

*Background:* Approximately 22 million Americans participate in paddling-related activities each year [1], contributing to the global paddlesports market, which was valued at $2.4 billion in 2020 [2]. Despite the sport's widespread participation, the integration of technology and machine learning (ML) into paddlesports remains limited. Furthermore, the high cost of private coaching and specialized equipment presents barriers for individuals seeking to enter the competitive domain. This preliminary study introduces a novel approach to paddling instruction through the development of an artificial intelligence (AI) based coaching application. The system leverages ML models trained on data from a small cohort of paddlers and delivers qualitative feedback via a large language model (LLM).

*Methods:* Participants were recruited in person through a collaboration with the New York University (NYU) Concrete Canoe Team. Data were collected over multiple paddling sessions in which participants wore Apple Watches and mobile phones (secured in sport straps) running motion-tracking software. Each participant completed two sessions: one involving intentionally suboptimal paddling form and another with corrected, optimal technique. The collected motion data were subjected to stroke segmentation analysis and feature extraction. Several machine learning models were developed using both raw and engineered features, including a Support Vector Classifier, Random Forest, Gradient Boosting Classifier, and Extremely Randomized Tree Classifier. A web-based interface was also developed to allow paddlers to assess their stroke quality.

*Findings:* A total of eight trials were conducted across four participants, yielding 66 stroke samples. Among the models evaluated, the extratree binary classification model demonstrated the highest performance, achieving an F score of 0.9496 under five-fold cross-validation for distinguishing between optimal and suboptimal strokes. The developed web interface effectively delivered both quantitative metrics and qualitative LLM-based feedback when applied to unseen data. It was also found that sensor placement closer to the wrists yielded better insights.

*Interpretation:* The results suggest that models trained on motion sensing data from smartwatches and smartphones, when integrated with comprehensive feedback systems, hold promise for developing a low-cost and accessible alternative to traditional paddlesports training. Despite the limited sample size, the findings underscore the practical utility of widely available consumer devices, such as Apple Watches and mobile phones, in capturing meaningful data to support stroke refinement and technique improvement.

**Research in context**

*Evidence before this study*

Previous approaches to assessing paddle stroke quality have often relied on specialized equipment or high-resolution video footage, which are typically costly, inaccessible, and require technical expertise. To explore prior works, we conducted a literature review on Google Scholar using the search terms "Canoe paddling," "Motion tracking," "Motion sensor placement," "Wearable sensors," and "Machine Learning."

Several studies have explored the use of wearable technology for paddling analysis. Liu et al. [3] employed wearable inertial measurement units (IMUs) to capture motion data, applied gradient descent for sensor fusion, and used machine learning for stroke phase segmentation. Their findings successfully differentiated stroke patterns between novices and experienced coaches, identifying key indicators such as stroke length, stroke rate, stroke rate variability, propulsion-to-recovery ratio, and rhythm. However, while effective, this approach required specialized and water-sensitive equipment and lacked a user-facing feedback mechanism capable of delivering intuitive, actionable insights. Other promising studies have used miniature inertial devices. McDonnell et al. [4] mounted sensors on kayaks and paddles to capture stroke duration and peak metrics from data streams, while Gomes et al. [5] used paddle-mounted sensors to detect pauses between strokes. However, such paddle-mounted approaches focus on equipment motion rather than that of the paddler, overlooking key aspects of muscle and body movement.

Video-based methods, such as those proposed by Sánchez et al. [6] and Tay et al. [7], have also produced valuable insights. Nevertheless, their reliance on high-resolution cameras, equipment operators, and unobstructed filming conditions limits their real-world applicability compared to wearable-based solutions.

In assessing the optimal placement of wearable sensors, we prioritized body-mounted configurations over paddle-mounted ones to better capture paddler biomechanics. Insights from adjacent domains were also considered. Teng et al. [8] studied sensor positioning for fall detection and identified the torso and waist as critical locations. Similarly, Urukalo et al. [9] highlighted the torso, upper arms, shoulders, and wrists as ideal for motion sensing. We modeled our study design based on conclusions from these studies.

Despite advances in these works, few studies have explored the use of readily available consumer devices such as smartwatches and smartphones for paddling analysis, leaving a notable gap in the development of accessible and cost-effective solutions.

*Added value of this study*

This study addresses limitations of prior research by employing smartphone and smartwatch applications to analyze motion data and provide both quantitative and qualitative feedback. By using a free app on widely available devices, the approach enhances accessibility to automated paddling analysis tools. Our models classifying stroke quality demonstrated strong predictive performance on unseen data. Feature importance analysis identified watch-recorded accelerations as key indicators of paddler proficiency. Additionally, the study presents a multimodal feedback system integrating dot plots, pie charts, and large language model (LLM)-generated qualitative insights based on visualized features.

*Implications of all available evidence*

This preliminary pilot study demonstrates the potential of smartwatches and mobile phones to deliver accessible, accurate, and cost-effective paddling stroke assessments through mobile and web applications. It also highlights the value of multimodal feedback, including outputs generated by large language models (LLMs). Sensor placement on the upper arms and wrists appears sufficient for capturing meaningful motion data. Overall, the findings support the feasibility of an ML-based training system for paddlers in real-world contexts, where affordability, accessibility, and precision are essential.

**Introduction**

Paddling is central to several popular sports, including rowing, canoeing, and kayaking, with tens of millions of participants worldwide engaging annually for both leisure and competition (1). Stroke technique is fundamental to success in paddlesports, prompting substantial investment in coaching by both competitive teams and novices. However, coaching costs, ranging from hundreds to thousands of dollars per year (10), pose a significant financial barrier, particularly for beginners. Over 37 million individuals participate in paddlesports annually (11), and the market, valued at $2.4 billion in 2020, (2) grew to $4.5 billion by 2023 and is projected to reach $7.8 billion by 2032 (12). Despite this growth, no widely accessible alternative to traditional coaching currently exists. Research has explored the potential of various sensors, machine learning techniques, and sensor placements to enhance paddling analysis.

Smartwatches are widely used for activity tracking, boasting 455 million users globally as of 2025 (14, 15) and adoption by 91% of Olympic athletes (16). Yet, training still relies primarily on human coaches, which can introduce subjectivity and risk of injury. The ubiquity of smartwatches and smartphones presents a promising platform for scalable, accessible paddling coaching applications.

Previous research on technology-driven paddling coaching has primarily focused on motion tracking using inertial sensors and video analysis (3-9). Inertial measurement units, piezoelectric sensors, and optical fiber sensors placed on the arms and waist have effectively captured paddling-related movement patterns, enabling differentiation of stroke quality (3-5, 17-20). However, these approaches often rely on specialized, costly equipment, limiting widespread adoption. Video-based methods have also been investigated (6-7, 21) but face practical constraints. The deployment of sensing technologies on widely available devices such as smartwatches and smartphones, coupled with feedback via web interfaces, remains relatively underexplored. Sensor placement on the body, particularly the waist, upper arms, and wrists, has demonstrated greater effectiveness than paddle-mounted sensors (3, 8, 9), consistent with findings in fall detection and dual-arm manipulation studies. While some prior work has examined mobile and web interfaces (21), these systems have mainly highlighted AI detection capabilities rather than delivering personalized feedback and actionable guidance to paddlers.

This study addresses key challenges in AI-driven sports coaching for paddlesports by leveraging motion data from mobile phones and smartwatches. Participants completed 3-minute paddling sessions kneeling beside a pool, performing both suboptimal and optimal stroke forms. We extracted and summarized features from the data, followed by feature selection to develop Support Vector Machine, Random Forest, Gradient Boosting Functions, and Extremely Randomized tree models. The final model was deployed as a standalone REST API, accessible via a user-friendly web interface that allows paddlers to upload smartwatch and mobile phone data and receive feedback and actionable insights in multiple formats.

**Methods**

Data were collected using custom mobile and smartwatch applications, with participants wearing a watch on each wrist and a phone secured to the upper arm using a wrist strap. Machine learning models were trained on features extracted from the recorded motion data. A web-based interface was developed to allow users to upload paddling data and receive detailed, user-friendly feedback. **Figure 1** illustrates the complete workflow, from data acquisition to model deployment and result delivery.

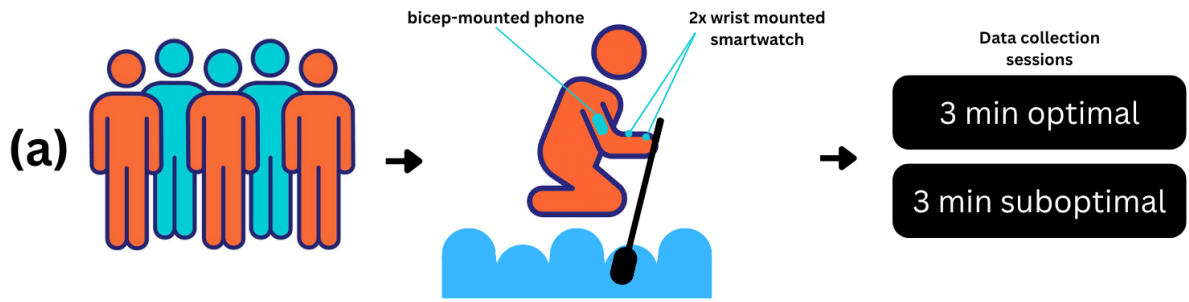

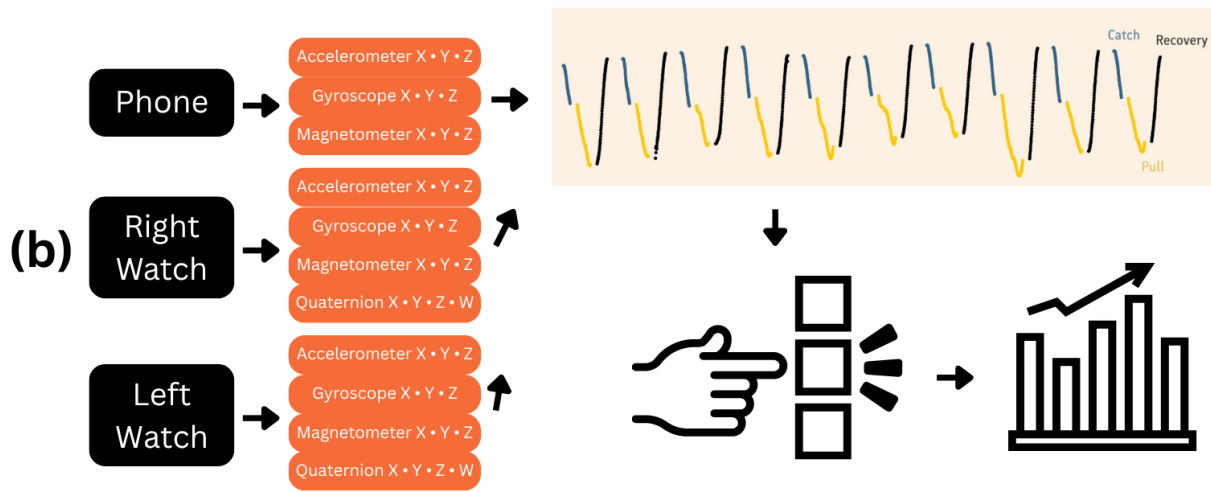

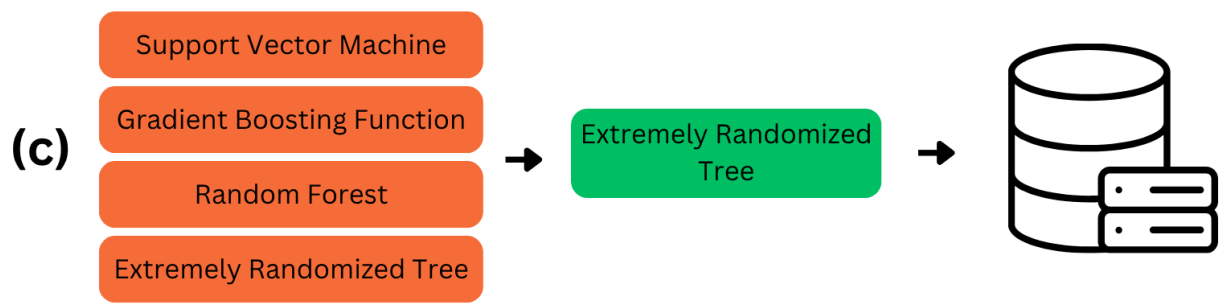

**Figure 1: The study workflow from data collection to model training and user-facing results.** (a) Participants completed the data collection process wearing a phone and smartwatches and paddled for predetermined periods. (b) Features were extracted and used to segment strokes. High-

performing features were then selected and converted to summaries for each stroke. (c) Models were evaluated, and the highest performing was selected and deployed.

*Participant Recruitment & Data Collection*

Participants were recruited in person through collaboration with the NYU Concrete Canoe Team, representing both experienced and novice paddlers. A dedicated pool lane was reserved in coordination with NYU Athletics to ensure an uninterrupted environment for data collection (surrounding pool lanes had minimal activity). All participants were instructed to paddle exclusively on their right side, irrespective of hand dominance, to maintain consistency across trials.

Data were collected using two Apple Watch Series 5 devices, one mounted on each wrist, and an iPhone 11 secured in a running armband on the upper arm, as shown in **Figure 2**. Multiple mobile and watch applications were evaluated for functionality, accuracy, and usability **(Table 1)**. The Sensor Logger app was selected for the iPhone, and the HemiPhysio Data app for the Apple Watches. Both watches were mounted in their standard orientation facing the paddler **(Figure 3)** and secured tightly to ensure stability. The iPhone was positioned vertically for readability. All devices were activated simultaneously; minor timing discrepancies were mitigated by later processing using each device's independent timestamp logging.

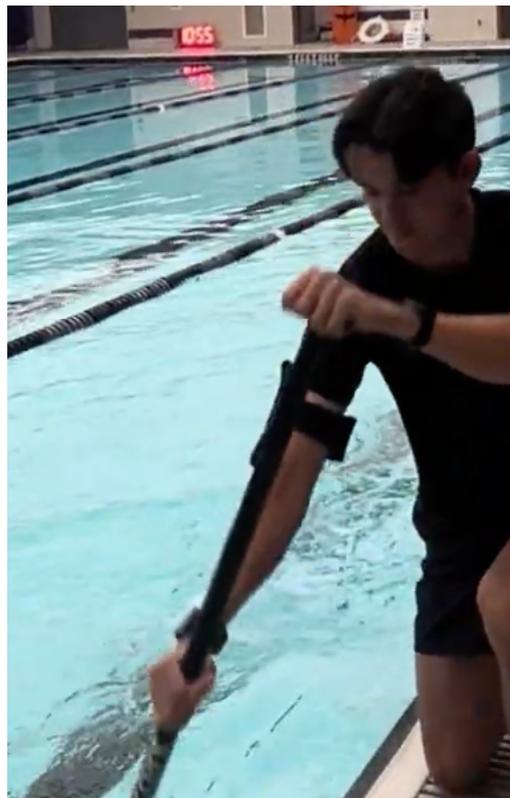

Figure 2: Positioning of Apple Watches and IPhone on person.

| App Name | Compatability | Benefits | Disadvantages |
| --- | --- | --- | --- |
| **Coremotion Data Recorder** | IPhone Apple Watch | Feature-rich | Dysfunctional on tested devices |
| **Sensor Data Recorder** | IPhone | 25 Hz sampling Records audio | No magnetometer reading Limit to 45 second recordings |
| **Sensor Logger** | IPhone Apple Watch | 100 Hz sampling Measures heart rate Records audio | Calibration necessary |
| **Hemiphysio Data** | Apple Watch | Feature rich | |
| **G-Field Recorder** | IPhone Apple Watch | Variable sampling rate | Export requires purchase |
| **Sensor-App** | IPhone Apple Watch | Has multitude of sensors | Cannot run all sensors at once |
| **Cumulocity IoT Sensor App** | IPhone | Has multitude of sensors | Cannot send CSVs |
| **Sensors Toolbox - Multitool** | IPhone | Has multitude of sensors | Must combine data manually |

**Table 1: Evaluation of several motion measurement apps.** Sensor Logger for IPhone and Hemiphysio Data for Apple Watch were ultimately chosen due to their combination of ease of use and adequate utility.

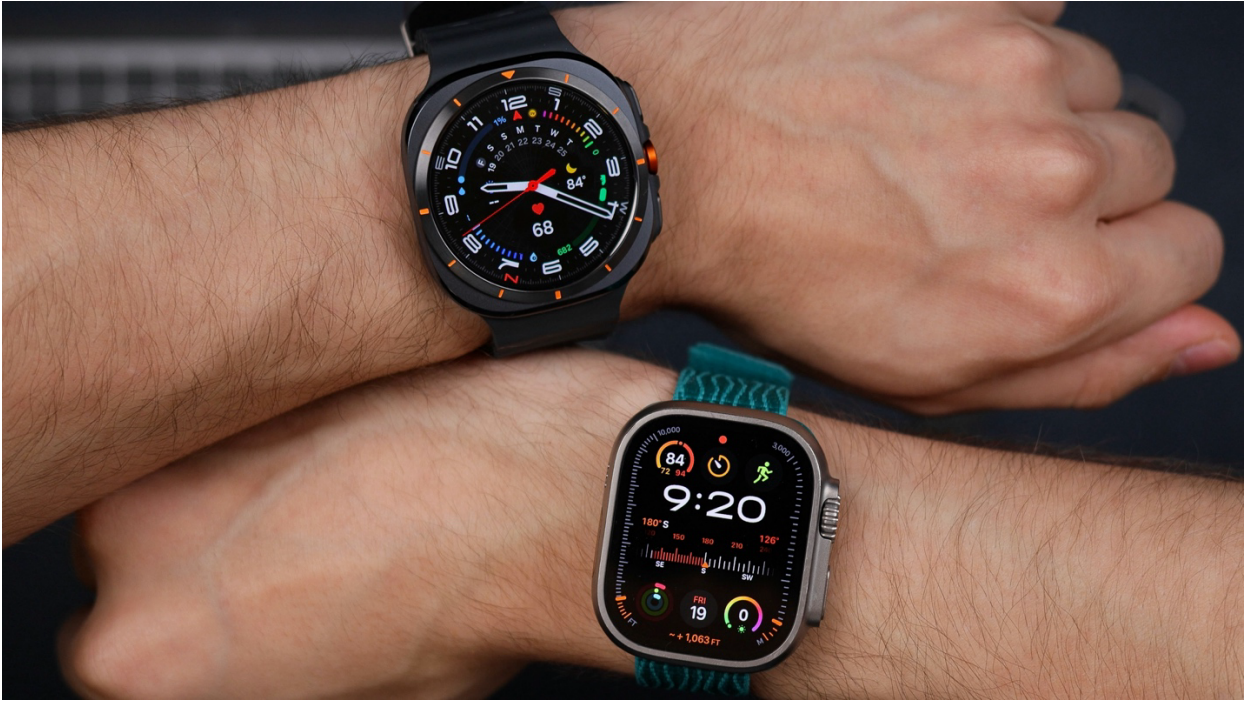

**Figure 3: Facing of Apple watches on either arm was in the natural direction.**

Each data collection session comprised two three-minute paddling trials per participant. The first involved intentionally suboptimal strokes, with participants instructed to introduce common paddling errors. The second focused on optimal paddling, during which participants aimed to maintain effective form and received guidance to refine their technique.

*Features*

We extracted 18 features from each Apple Watch and 9 from the iPhone. All devices recorded acceleration and rotational data along the X, Y, and Z axes. The Apple Watches additionally captured orientation (roll, pitch, yaw), gravity vectors (X, Y, Z), quaternion values (W, X, Y, Z), and processed acceleration. Among these, the quaternion stream yielded the most interpretable and informative signal. The iPhone also provided magnetometer data along the X, Y, and Z axes, complementing the shared motion data streams.

*Data Processing*

The raw sensor data required transformation into engineered features through a multi-step preprocessing pipeline: (1) time-series alignment to synchronize data streams across devices, (2) stroke segmentation to isolate individual strokes, (3) phase segmentation to identify stroke components (catch, pull, and recovery), (4) standardization of stroke duration, and (5) summarization to reduce feature dimensionality. This workflow is illustrated in **Figure 4**.

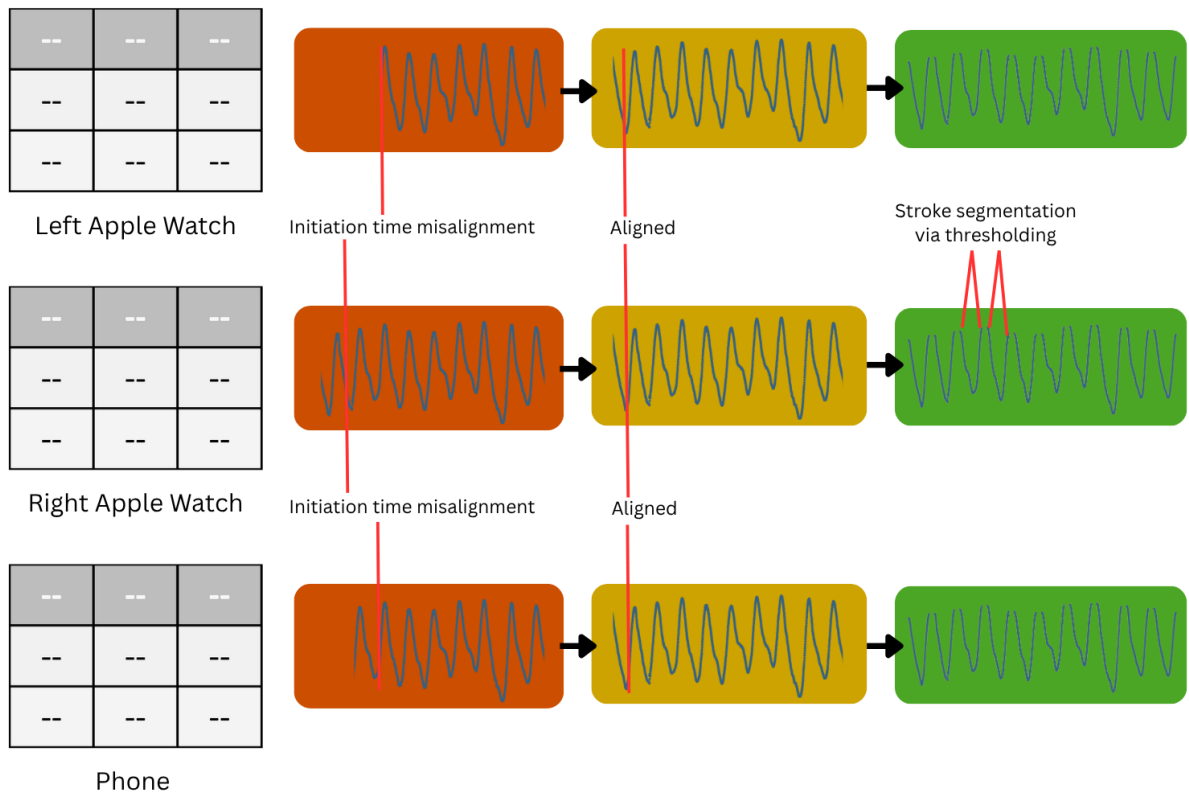

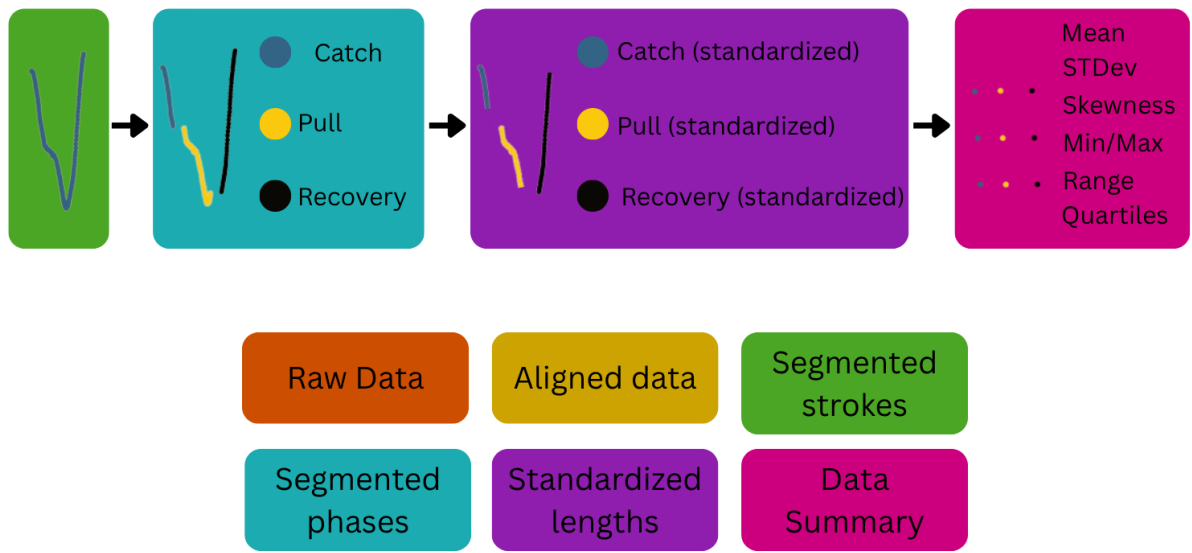

**Figure 4: Data processing from raw data to summary.** Misalignment in recording initiation time is rectified using timestamps, followed by stroke segmentation using thresholding. The

**segmented stroke is split into phases via segmentation, then standardized for length. Finally, summary statistics are extracted.**

*Alignment*

Simultaneous recording across all devices was not guaranteed due to human error and variations in sampling rates between devices and applications. However, each data file (two from the Apple Watches and three from the iPhone) contained timestamps in milliseconds or nanoseconds. This enabled construction of a unified dataset for each trial through temporal alignment of the five data files.

*Stroke Segmentation*

To enable stroke-level analysis, individual strokes were segmented using a threshold-based method. Across all trials, the X-component of the quaternion stream from the left Apple Watch consistently exhibited the clearest pattern and strongest alignment with video recordings, achieving peak values between strokes **(Figures 5 and 6)**. Feature graphs were synchronized with video footage to visually and quantitatively evaluate their correspondence with stroke initiation and termination. **Figure 7** illustrates the segmented X quaternion values from the left watch. Automated segmentation error was minimal, and visual inspection further validated its accuracy.

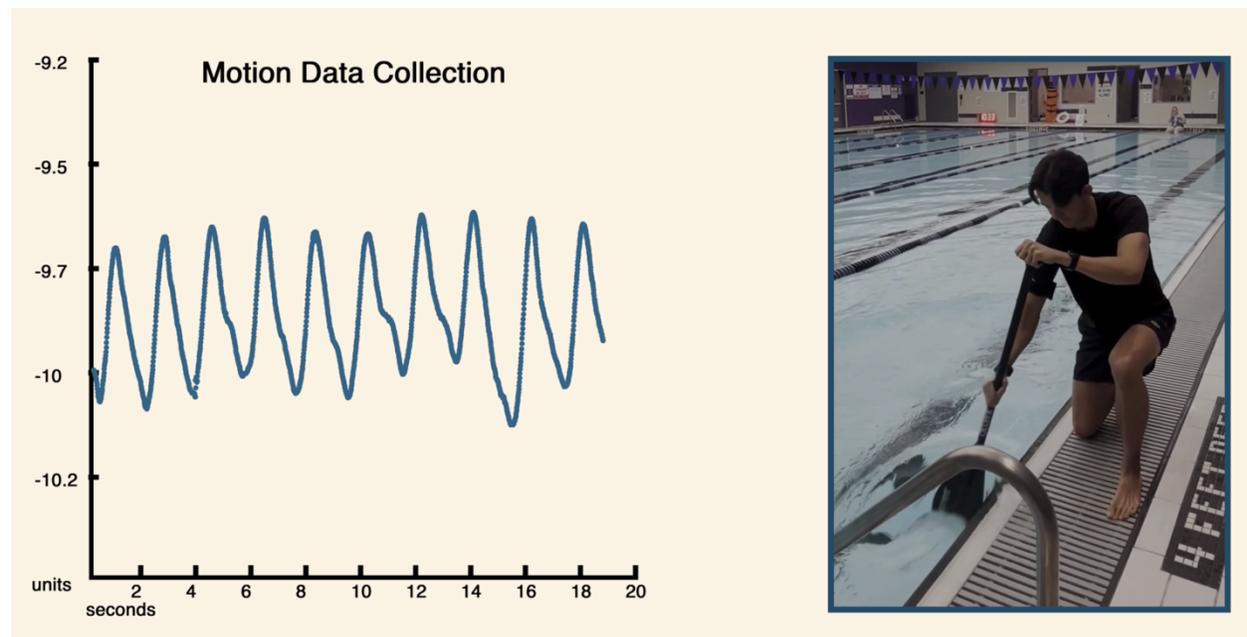

**Figure 5: Graphs of each feature were plotted in parallel to video footage.**

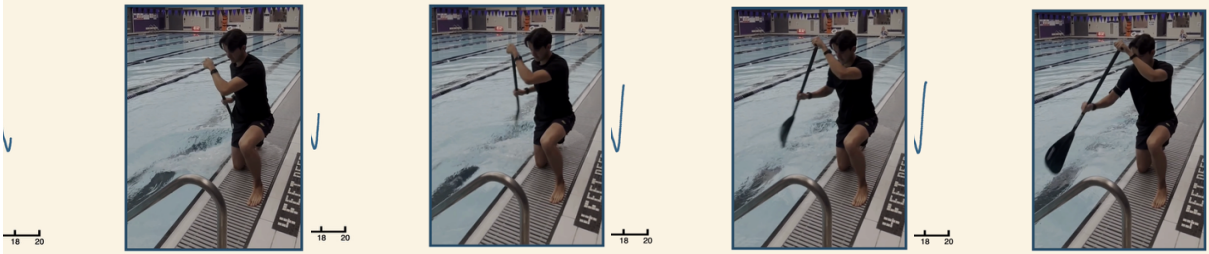

**Figure 6: A steep upward trend (as shown) in the Left Watch's X quaternion corresponded to the gap in strokes.**

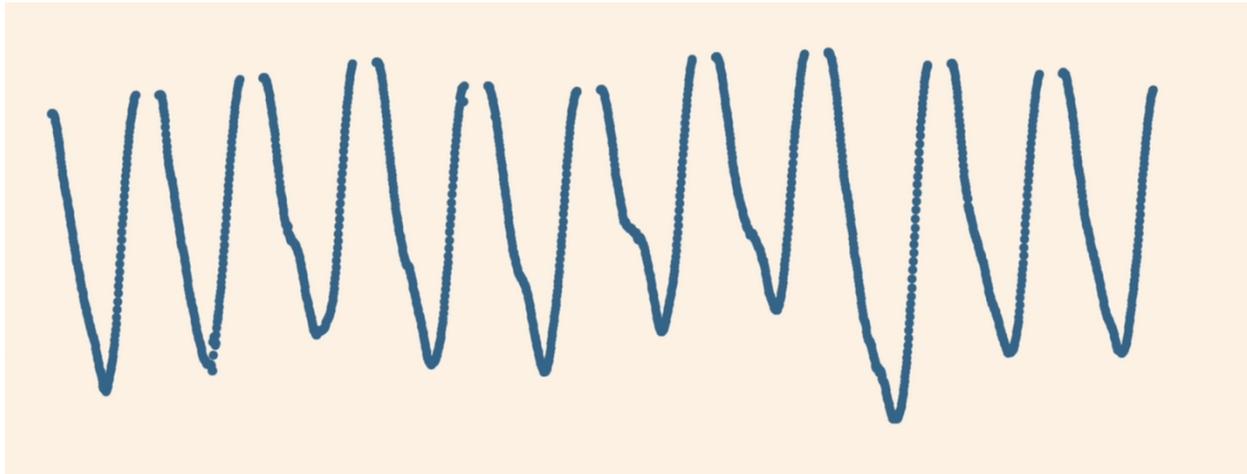

**Figure 7: Depiction of strokes after segmentation.**

*Stroke Phases Segmentation*

To allow for more actionable feedback, individual stroke phases were segmented and analyzed separately using a thresholding approach. Among all features, the W-component of the quaternion stream from the left Apple Watch was the most indicative of stroke phase transitions. As illustrated in **Figure 8**, consistent patterns corresponding to the Catch, Pull, and Recovery phases were observed. Feature graphs were evaluated alongside synchronized video footage to assess the accuracy of segmentation. **Figure 9** displays the W quaternion values following phase segmentation. Automated segmentation error was minimal, and visual validation confirmed its reliability.

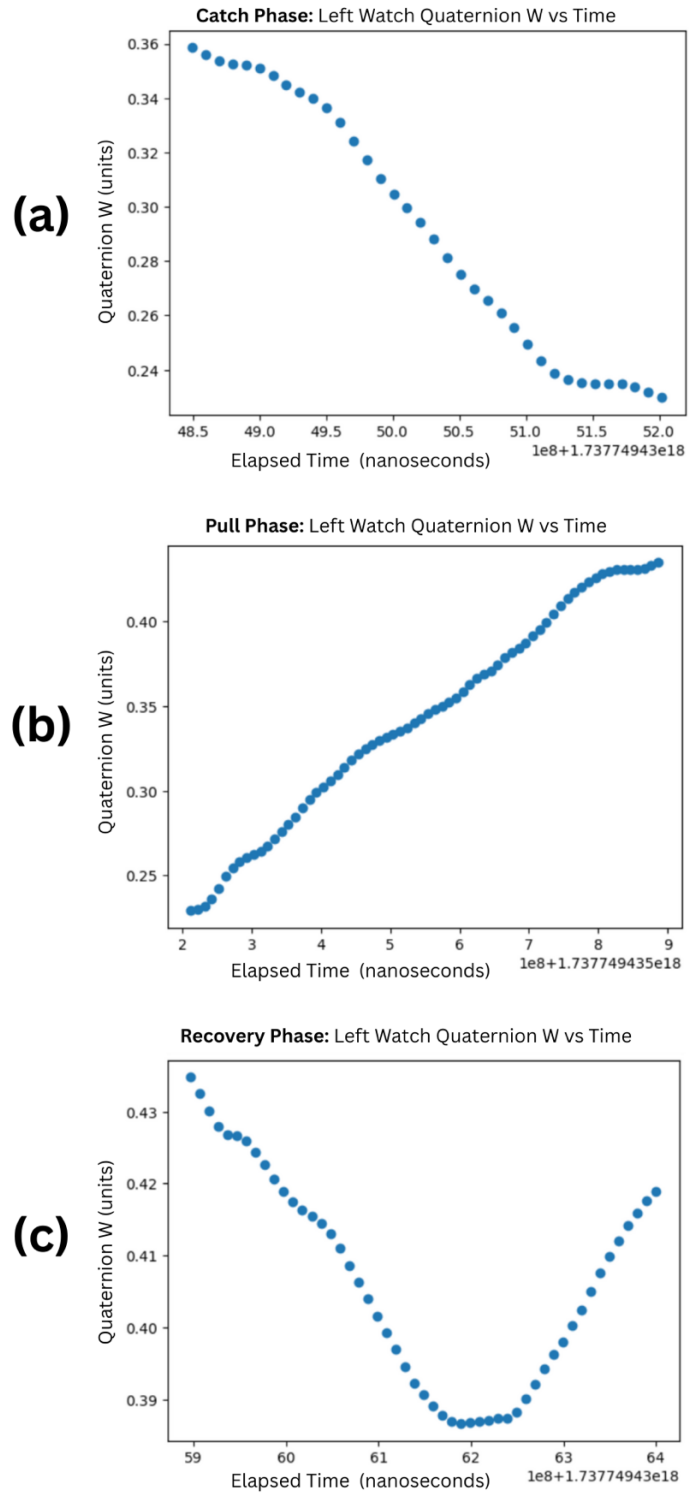

**Figure 8: Consistent trends in Left Watch Quaternion W values, seen after thresholding segmentation.** (a) a consistent downward trend can be observed in the catch phase, as the paddler rotates the paddler into the water. (b) an upward trend is indicated in the pull phase, as the paddler modifies the rotation to drive the stroke. (c) a parabolic path is seen in the recovery phase, as the paddler rotates the paddle out of the water.

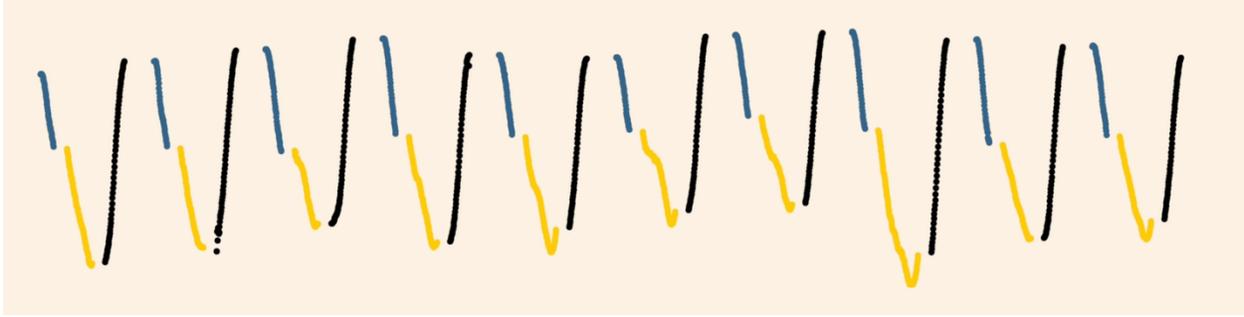

**Figure 9: Depiction of strokes after phase segmentation.**

*Standardization*

The thresholding-based segmentation process occasionally produced inaccurate stroke phase endpoints, with segmentation often initiating correctly but terminating imprecisely. To address this issue, each segmented stroke phase was truncated, retaining only the initial 40 frames for subsequent analysis. This approach not only reduced noise but also enabled preliminary model training on raw data (since the number of features would remain constant from stroke to stroke). However, the resulting model trained on raw data was not utilized in the final implementation.

*Data Summaries*

To reduce model input dimensionality while preserving essential stroke characteristics, data summarization was employed. Summary statistics included mean, skewness, standard deviation, minimum, maximum, range, 25th percentile, and 75th percentile. Transitioning from raw data to summarized features resulted in a marginal decrease in accuracy (<5%), rendering the summarized model a more efficient and practical choice. Data summarization also proved advantageous for elucidating feature importance analysis results.

*Stroke Selection*

As indicated before, the stroke segmentation process was susceptible to inaccuracies due to rest periods and signal artifacts caused by paddle contact with the pool wall. To ensure the inclusion of only well-segmented strokes, several quality control measures were implemented. These included evaluating the duration of each stroke phase, excluding strokes with phases deemed too short, and verifying the successful identification of all stroke phases, as illustrated in **Figure 10**.

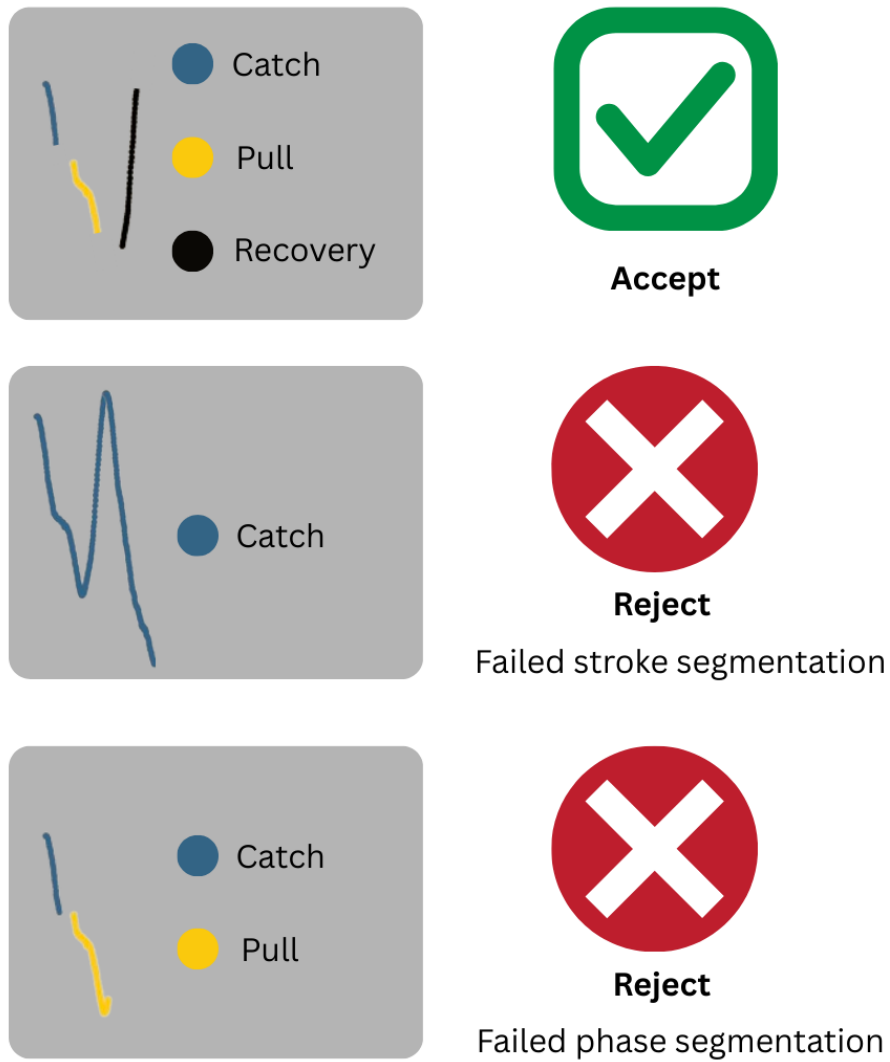

**Figure 10: Rejection of strokes that were segmented improperly**

*Model Development*

Through this process, we extracted 45 features for each stroke phase. Given that each paddling sample was labeled as optimal or suboptimal, these features directly mapped to a binary output. We developed four classifiers for each phase: a Support Vector Classifier (SVC) with an automatic gamma setting; a Random Forest Classifier with a maximum depth of two; a Gradient Boosting Classifier with 100 estimators and maximum depth of one; and an Extremely Randomized Tree Classifier with 100 estimators and maximum depth of one. All models were binary classifiers and validated using 5-fold cross-validation to prevent overfitting.

*User-Facing Interface*

To enhance accessibility for non-technical paddlers, we developed an intuitive web interface enabling users to upload data files and receive feedback. The pretrained model, stored as a Pickle file, was deployed via a REST API. The frontend, built with HTML, CSS, and JavaScript, processes uploaded data following the training pipeline: alignment, stroke segmentation, phase segmentation, standardization, and summarization. Processed data is sent to the model API through a GET request, which returns a stroke quality prediction (optimal or suboptimal). Results are presented to users through pie charts and qualitative feedback generated by an LLM via Deepseek. Deepseek is prompted with raw data streams, allowing the LLM to directly interface with comprehensive data. The overall workflow is illustrated in **Figure 11**.

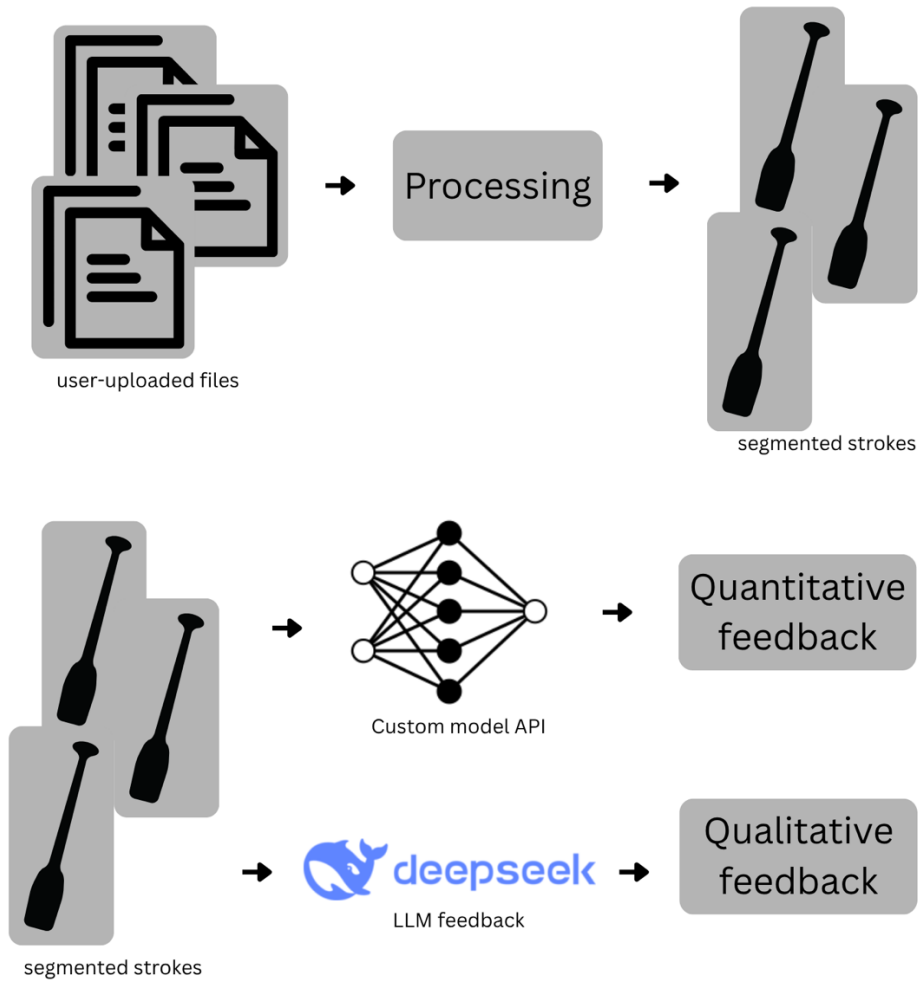

Figure 11: Workflow of the user-facing website

*User-Facing Graphs*

Side-by-side graphs comparing the user's stroke with a pre-selected optimal stroke were provided as feedback. Following segmentation, a segment of the user's stroke was automatically selected.

To exclude rest periods, only consecutive non-rejected strokes were considered, and the longest such sequence (capped at eight strokes) was displayed alongside the optimal sample.

## Results

*Dataset*

Four participants completed data collection on January 24, 2025, yielding 66 stroke samples. Among them, three were experienced paddlers and one was a novice. All participants were aged 18 to 23. Each completed two 3-minute trials: one optimal and one suboptimal paddling session.

*Binary Classification Models*

Our primary analysis evaluated the effectiveness of various models in classifying stroke phases as optimal or suboptimal using features from all devices. The models tested included Support Vector Classifier (SVC), Random Forest (RF), Gradient Boosting Function (GBF), and Extremely Randomized Tree (Extratree) classifiers. Key metrics, including means and standard deviations, are presented in **Table 2** and **Figure 12**. For the Catch phase, both SVC and Extratree models demonstrated superior performance, achieving identical accuracies of 0.9545 ± 0.0444, sensitivities of 0.9231 ± 0.0568, negative predictive values (NPV) of 0.9000 ± 0.0640, and F scores of 0.9600 ± 0.0418. During the Pull phase, RF, GBF, and Extratree models achieved similar accuracies (0.9545 ± 0.0444), with GBF exhibiting the highest sensitivity (0.9231 ± 0.0568) and F score (0.9600 ± 0.0418), while RF had the highest NPV (0.9375 ± 0.0516). In the Recovery phase, the Extratree model outperformed others, with accuracy of 0.9545 ± 0.0444, sensitivity of 0.9000 ± 0.0640, NPV of 0.9231 ± 0.0568, and F score of 0.9474 ± 0.0476. Overall, the Extratree model exhibited the best performance across phases. Models generally performed best in the Recovery phase, followed by the Pull and Catch phases.

| Stroke Phase | Model | Accuracy Mean ± SE | Sensitivity Mean ± SE | Specificity Mean ± SE | PPV Mean ± SE | NPV Mean ± SE | F Score Mean ± SE |
|---|---|---|---|---|---|---|---|
| **Catch** | SVC | 0.9545 ± 0.0444 | 0.9231 ± 0.0568 | 1.0000 ± 0.0000 | 1.0000 ± 0.0000 | 0.9000 ± 0.0640 | 0.9600 ± 0.0418 |
| | RF | 0.9091 ± 0.0613 | 0.8462 ± 0.0769 | 1.0000 ± 0.0000 | 1.0000 ± 0.0000 | 0.8182 ± 0.0822 | 0.9167 ± 0.0589 |
| | GBF | 0.8636 ± 0.0732 | 0.7692 ± 0.0898 | 1.0000 ± 0.0000 | 1.0000 ± 0.0000 | 0.7500 ± 0.0923 | 0.8696 ± 0.0718 |
| | Extratree | 0.9545 ± 0.0444 | 0.9231 ± 0.0568 | 1.0000 ± 0.0000 | 1.0000 ± 0.0000 | 0.9000 ± 0.0640 | 0.9600 ± 0.0418 |
| **Pull** | SVC | 0.8636 ± 0.0732 | 0.7692 ± 0.0898 | 1.0000 ± 0.0000 | 1.0000 ± 0.0000 | 0.7500 ± 0.0923 | 0.8696 ± 0.0718 |

|  | | | | | | | | | | | | |
|---|---|---|---|---|---|---|---|---|---|---|---|---|
|  | RF | 0.9545 | ± | 0.8571 | ± | 1.0000 | ± | 1.0000 | ± | 0.9375 | ± | 0.9231 | ± |
|  |  | 0.0444 |  | 0.0746 |  | 0.0000 |  | 0.0000 |  | 0.0516 |  | 0.0568 |  |
|  | GBF | 0.9545 | ± | 0.9231 | ± | 1.0000 | ± | 1.0000 | ± | 0.9000 | ± | 0.9600 | ± |
|  |  | 0.0444 |  | 0.0568 |  | 0.0000 |  | 0.0000 |  | 0.0640 |  | 0.0418 |  |
|  | Extratree | 0.9545 | ± | 0.8889 | ± | 1.0000 | ± | 1.0000 | ± | 0.9286 | ± | 0.9412 | ± |
|  |  | 0.0444 |  | 0.0670 |  | 0.0000 |  | 0.0000 |  | 0.0549 |  | 0.0502 |  |
| Recovery | SVC | 0.9545 | ± | 0.8889 | ± | 1.0000 | ± | 1.0000 | ± | 0.9286 | ± | 0.9412 | ± |
|  |  | 0.0444 |  | 0.0670 |  | 0.0000 |  | 0.0000 |  | 0.0549 |  | 0.0502 |  |
|  | RF | 0.9545 | ± | 0.8889 | ± | 1.0000 | ± | 1.0000 | ± | 0.9286 | ± | 0.9412 | ± |
|  |  | 0.0444 |  | 0.0670 |  | 0.0000 |  | 0.0000 |  | 0.0549 |  | 0.0502 |  |
|  | GBF | 0.9091 | ± | 0.8889 | ± | 0.9231 | ± | 0.8889 | ± | 0.9231 | ± | 0.8889 | ± |
|  |  | 0.0613 |  | 0.0670 |  | 0.0568 |  | 0.0670 |  | 0.0568 |  | 0.0670 |  |
|  | Extratree | 0.9545 | ± | 0.9000 | ± | 1.0000 | ± | 1.0000 | ± | 0.9231 | ± | 0.9474 | ± |
|  |  | 0.0444 |  | 0.0640 |  | 0.0000 |  | 0.0000 |  | 0.0568 |  | 0.0476 |  |

**Table 2: Performance metrics for models trained and tested on optimal and suboptimal strokes for the Catch, Pull, and Recovery phases.**

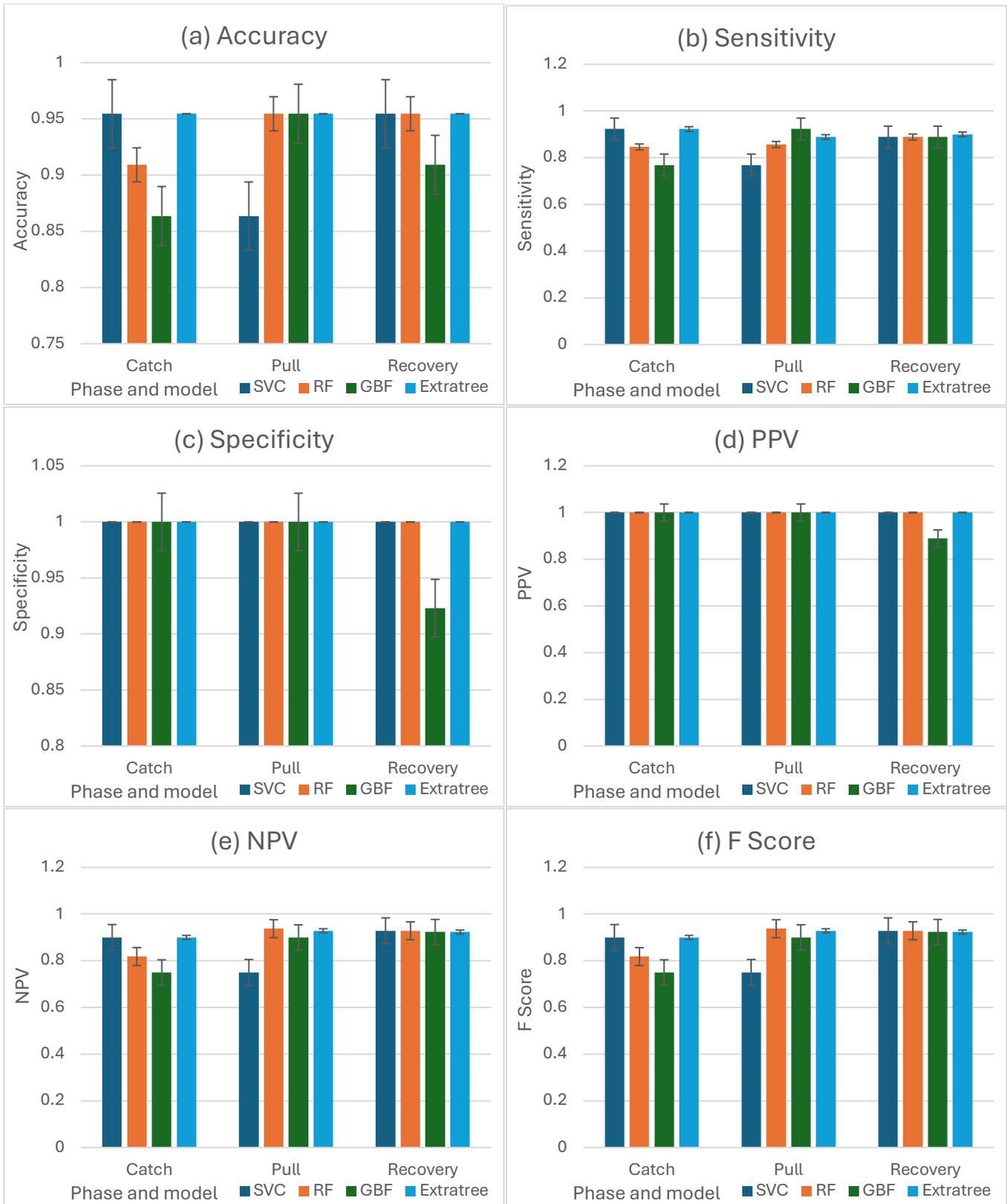

Figure 12: The performance metrics for models; Sensitivity (a), Specificity (b), PPV (c), NPV (d), Accuracy (e), and F Score (f). The error bars represent standard error values.

*Device positioning*

Our secondary analysis evaluated data quality based on device placement, specifically, the hand on the paddle's upper handle, the paddle stem, and the bicep. We used an Extratree classifier, the top-performing model across stroke phases. For the Catch phase, the left wrist-mounted device achieved the highest performance, with accuracy of 0.9545 ± 0.0444, sensitivity of 0.9000 ± 0.0640, NPV of 0.9231 ± 0.0568, and F score of 0.9474 ± 0.0476. The right wrist-mounted device performed similarly for the Catch phase, with accuracy of 0.9545 ± 0.0444, sensitivity of 0.9091 ± 0.0613, NPV of 0.9167 ± 0.0589, and F score of 0.9524 ± 0.0454. For the Recovery phase, the right wrist-mounted device yielded the best results, with accuracy of 0.9545 ± 0.0444, sensitivity of 0.9000 ± 0.0640, NPV of 0.9231 ± 0.0568, and F score of 0.9474 ± 0.0476. Overall, the left wrist-mounted device produced the best model performance, closely followed by the right wrist-mounted device. Detailed results are presented in **Table 3** and **Figure 13**.

| Stroke Phase | Device Location | Accuracy Mean ± SE | Sensitivity Mean ± SE | Specificity Mean ± SE | PPV Mean ± SE | NPV Mean ± SE | F Score Mean ± SE |
|---|---|---|---|---|---|---|---|
| Catch | Right Bicep | 0.9091 ± 0.0613 | 0.7778 ± 0.0886 | 1.0000 ± 0.0000 | 1.0000 ± 0.0000 | 0.8667 ± 0.0725 | 0.8750 ± 0.0705 |
| | Left Wrist | 0.9545 ± 0.0444 | 0.9000 ± 0.0640 | 1.0000 ± 0.0000 | 1.0000 ± 0.0000 | 0.9231 ± 0.0568 | 0.9474 ± 0.0476 |
| | Right Wrist | 0.9091 ± 0.0613 | 0.8462 ± 0.0769 | 1.0000 ± 0.0000 | 1.0000 ± 0.0000 | 0.8182 ± 0.0822 | 0.9167 ± 0.0589 |
| Pull | Right Bicep | 0.8182 ± 0.0822 | 1.0000 ± 0.0000 | 0.7647 ± 0.0904 | 0.5556 ± 0.1059 | 1.0000 ± 0.0000 | 0.7143 ± 0.0963 |
| | Left Wrist | 0.9545 ± 0.0444 | 0.9000 ± 0.0640 | 1.0000 ± 0.0000 | 1.0000 ± 0.0000 | 0.9231 ± 0.0568 | 0.9474 ± 0.0476 |
| | Right Wrist | 0.9545 ± 0.0444 | 0.9091 ± 0.0613 | 1.0000 ± 0.0000 | 1.0000 ± 0.0000 | 0.9167 ± 0.0589 | 0.9524 ± 0.0454 |
| Recovery | Right Bicep | 0.9091 ± 0.0613 | 0.7778 ± 0.0886 | 1.0000 ± 0.0000 | 1.0000 ± 0.0000 | 0.8667 ± 0.0725 | 0.8750 ± 0.0705 |
| | Left Wrist | 0.9545 ± 0.0444 | 1.0000 ± 0.0000 | 0.9412 ± 0.0502 | 0.8333 ± 0.0795 | 1.0000 ± 0.0000 | 0.9091 ± 0.0613 |
| | Right Wrist | 0.9545 ± 0.0444 | 0.9000 ± 0.0640 | 1.0000 ± 0.0000 | 1.0000 ± 0.0000 | 0.9231 ± 0.0568 | 0.9474 ± 0.0476 |

**Table 3: Performance metrics for data sources (devices) used for model training for the Catch, Pull, and Recovery phases.**

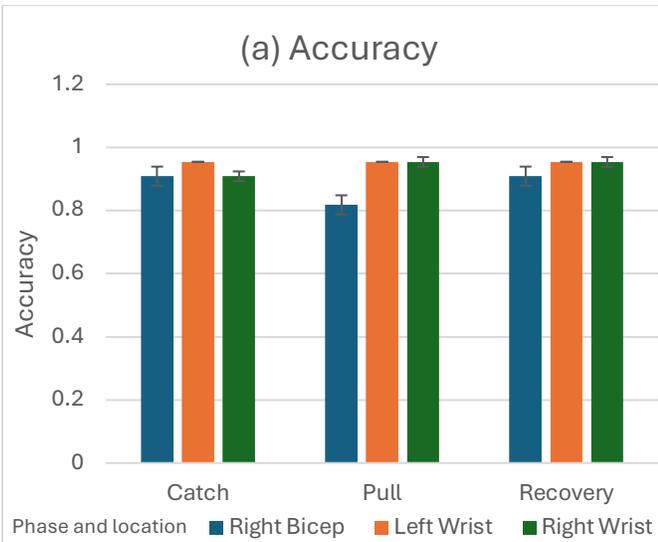
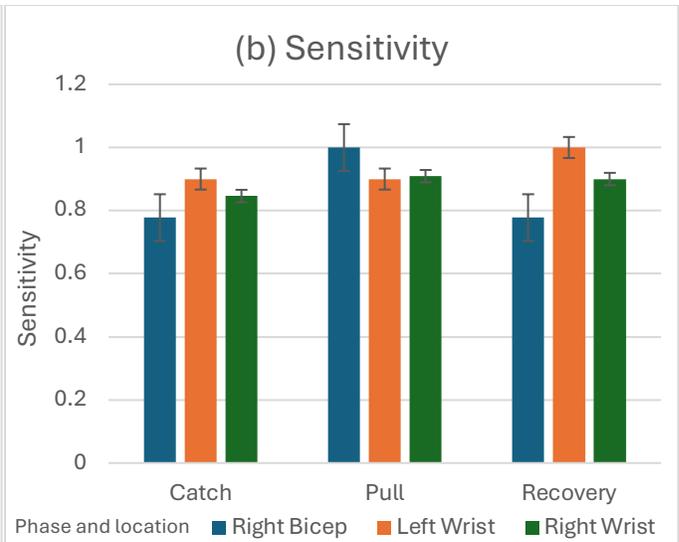
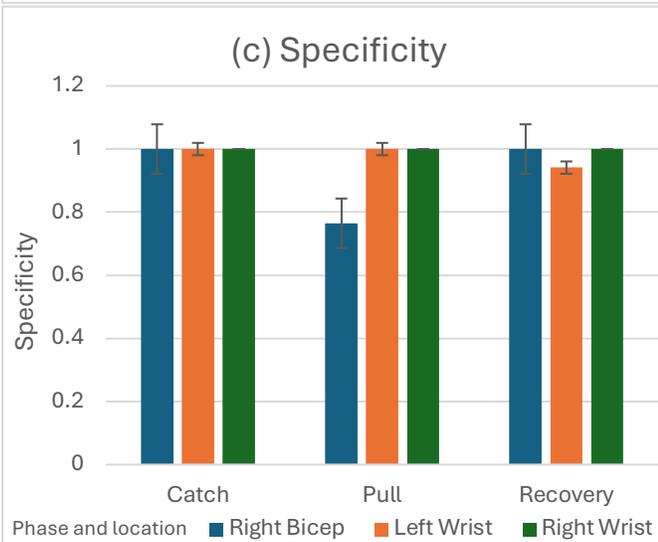
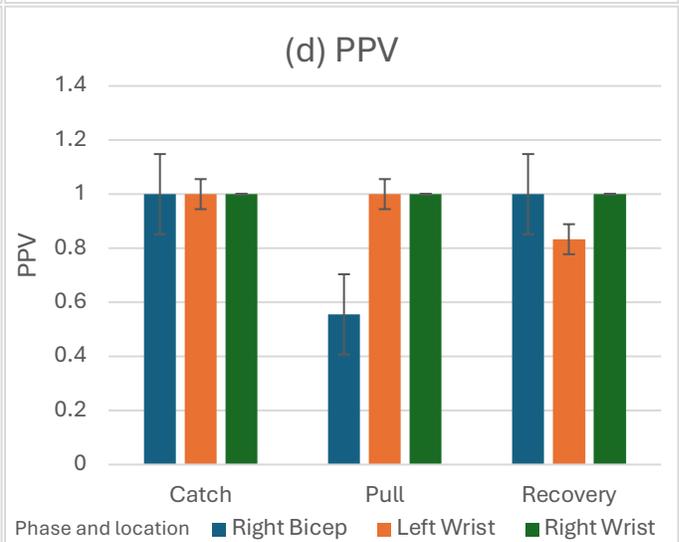
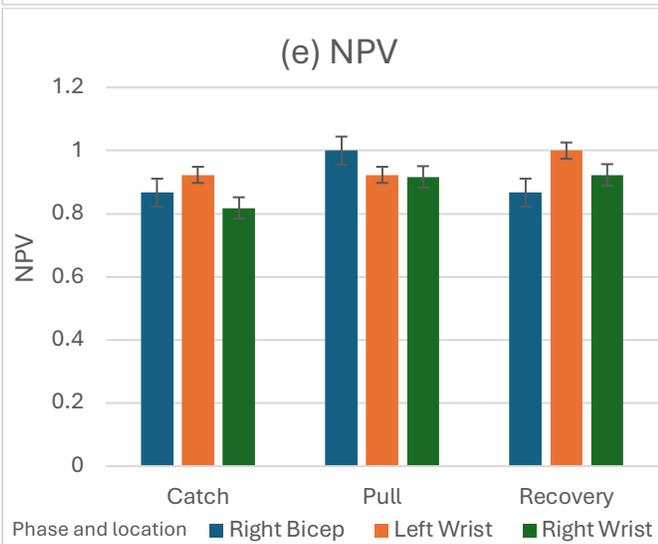
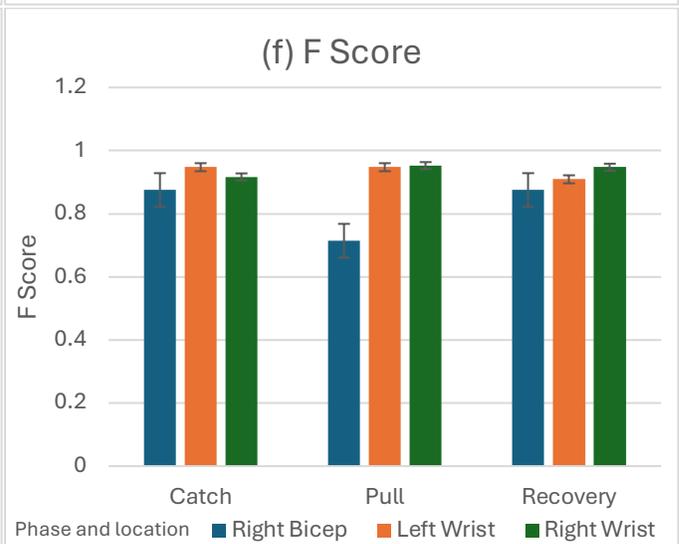

**Figure 13:** The performance metrics for Extratree models trained on data from specific devices; Sensitivity (a), Specificity (b), PPV (c), NPV (d), Accuracy (e), and F Score (f). The error bars represent standard error values.

*Feature Importance*

Permutation feature importance analysis was conducted to assess the significance of various data streams. The data was categorized into seven groups: phone accelerometer, left watch accelerometer, right watch accelerometer, phone gyroscope, left watch rotation, right watch rotation, and phone magnetometer. For each category, all axes (X, Y, Z) were analyzed alongside summary statistics including mean, skewness, standard deviation, minimum, maximum, range, first quartile, and third quartile. As illustrated in **Figure 14**, several key features were identified, notably phone accelerometer Z skewness, phone accelerometer Z first quartile, left watch accelerometer Y first and third quartiles, phone gyroscope Z maximum, and phone magnetometer X maximum. However, this feature importance analysis yielded largely inconclusive results, with few apparent meaningful trends.

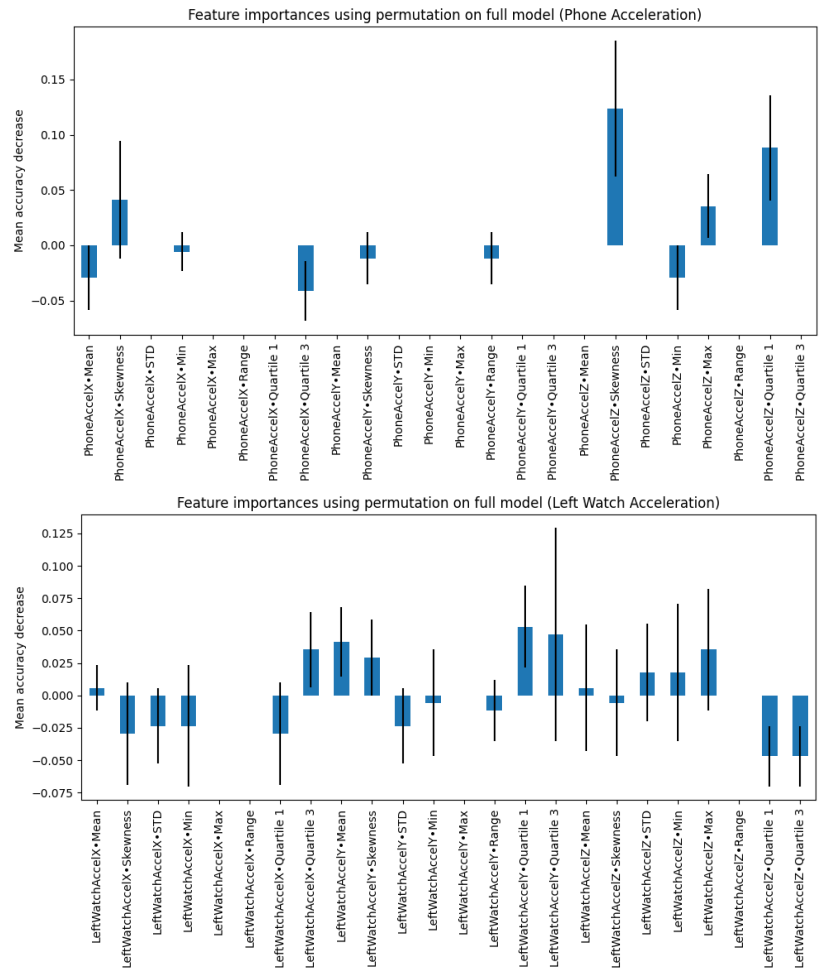

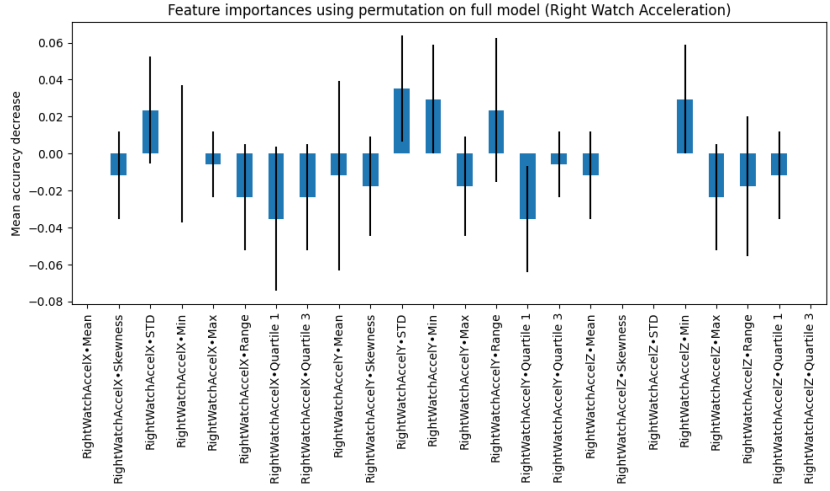
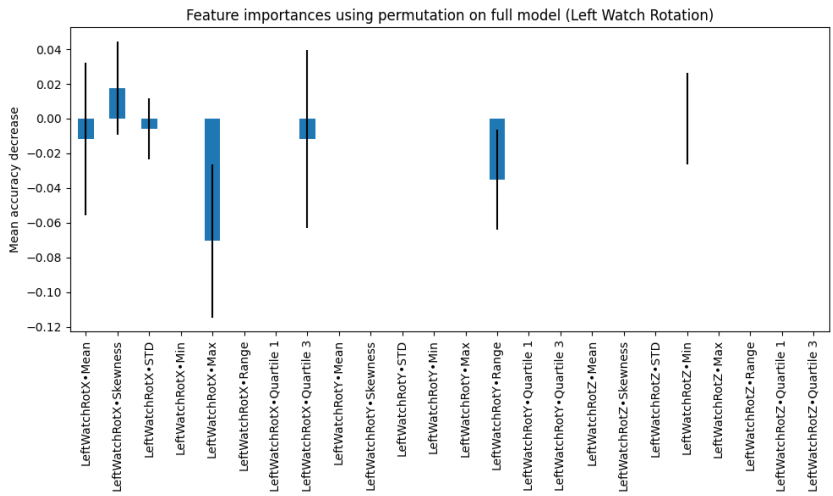
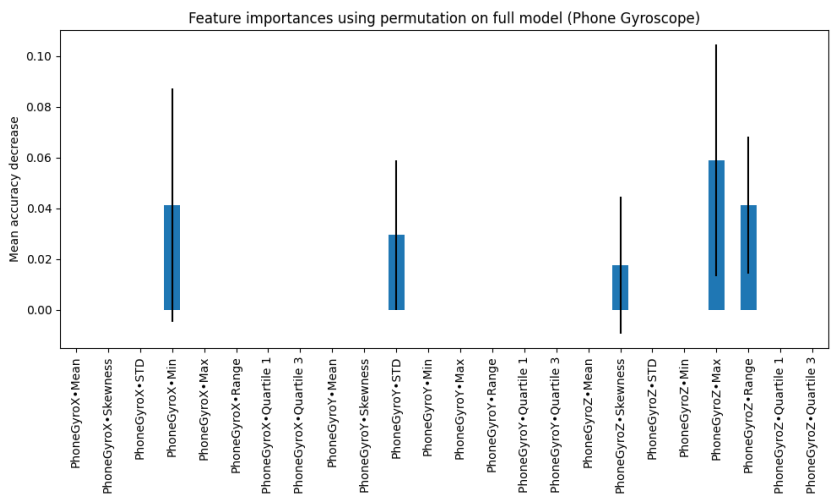

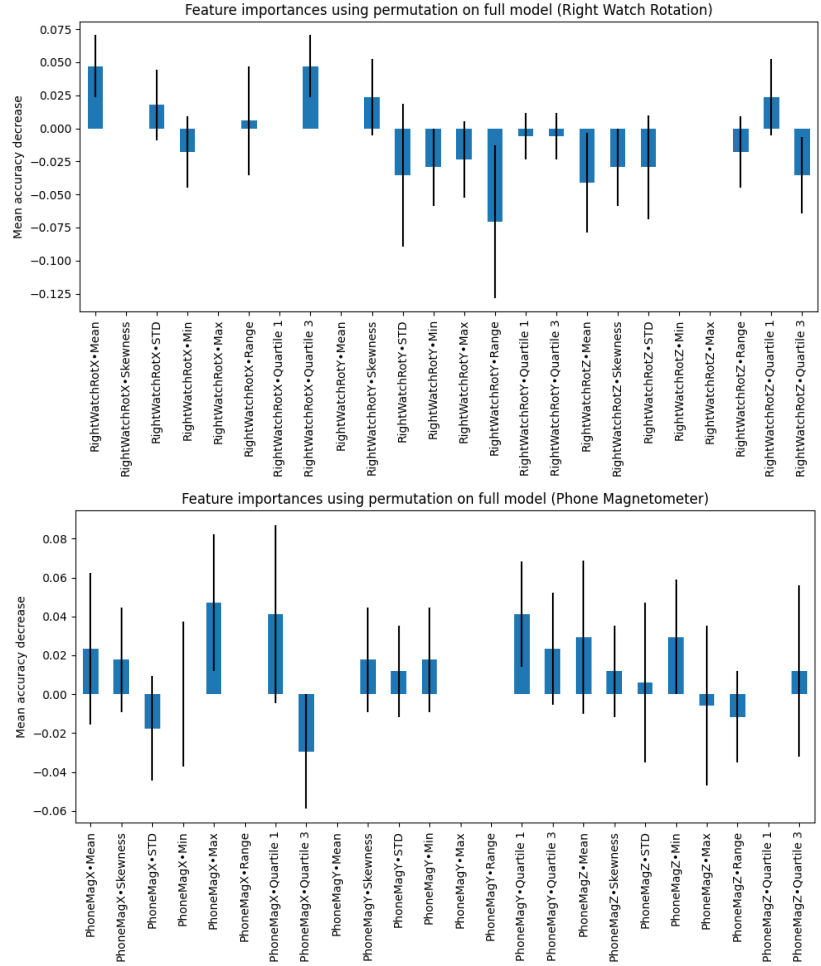

**Figure 14: Feature Importances using permutations. No notable trends were observed in feature importances.**

*Anomaly Detection*

Anomalies such as water disturbance during the catch phase, contact with the boat or pool surfaces, and improper recovery technique may indicate suboptimal strokes. These anomalies are reflected in the collected data, prompting the evaluation of anomaly detection models. For the catch phase, the Isolation Forest model demonstrated strong performance, achieving an accuracy of 0.8889 ± 0.0524, sensitivity of 0.9231 ± 0.0444, specificity of 0.8000 ± 0.0667, PPV of 0.9231 ± 0.0444, negative NPV of 0.8000 ± 0.0667, and F score of 0.9231 ± 0.0444. However, its performance remained slightly inferior to the SVC and Extratree models, which showed higher accuracy. For the pull phase, the One-Class SVM achieved the highest accuracy (0.8333 ± 0.0621) and F score (0.8966 ± 0.0508) among anomaly detectors, but still underperformed compared to the best classification model, the Gradient Boosting Classifier. In the recovery phase, the Isolation Forest performed best among anomaly detectors with accuracy of 0.8611 ± 0.0576 and F score of 0.9123 ± 0.0471, yet it lagged behind the Extratree model. Overall, anomaly detection models underperformed relative to traditional binary classifiers, as summarized in **Table 4** and **Figure 15**.

| Stroke Phase | Model | Accuracy Mean ± SE | Sensitivity Mean ± SE | Specificity Mean ± SE | PPV Mean ± SE | NPV Mean ± SE | F Score Mean ± SE |
|---|---|---|---|---|---|---|---|
| Catch | Isolation Forest | 0.8889 ± 0.0524 | 0.9231 ± 0.0444 | 0.8000 ± 0.0667 | 0.9231 ± 0.0444 | 0.8000 ± 0.0667 | 0.9231 ± 0.0444 |
| | One-Class SVM | 0.8056 ± 0.0660 | 1.0000 ± 0.0000 | 0.3000 ± 0.0764 | 0.7879 ± 0.0681 | 1.0000 ± 0.0000 | 0.8814 ± 0.0539 |
| Pull | Isolation Forest | 0.7222 ± 0.0769 | 0.6923 ± 0.0769 | 0.8000 ± 0.0667 | 0.9000 ± 0.0500 | 0.5000 ± 0.0833 | 0.7826 ± 0.0687 |
| | One-Class SVM | 0.8333 ± 0.0621 | 1.0000 ± 0.0000 | 0.4000 ± 0.0816 | 0.8125 ± 0.0651 | 1.0000 ± 0.0000 | 0.8966 ± 0.0508 |
| Recovery | Isolation Forest | 0.8611 ± 0.0576 | 1.0000 ± 0.0000 | 0.5000 ± 0.0833 | 0.8387 ± 0.0613 | 1.0000 ± 0.0000 | 0.9123 ± 0.0471 |
| | One-Class SVM | 0.8333 ± 0.0621 | 1.0000 ± 0.0000 | 0.4000 ± 0.0816 | 0.8125 ± 0.0651 | 1.0000 ± 0.0000 | 0.8966 ± 0.0508 |

**Table 4: Performance metrics for anomaly detection models for Catch, Pull, and Recovery phases.**

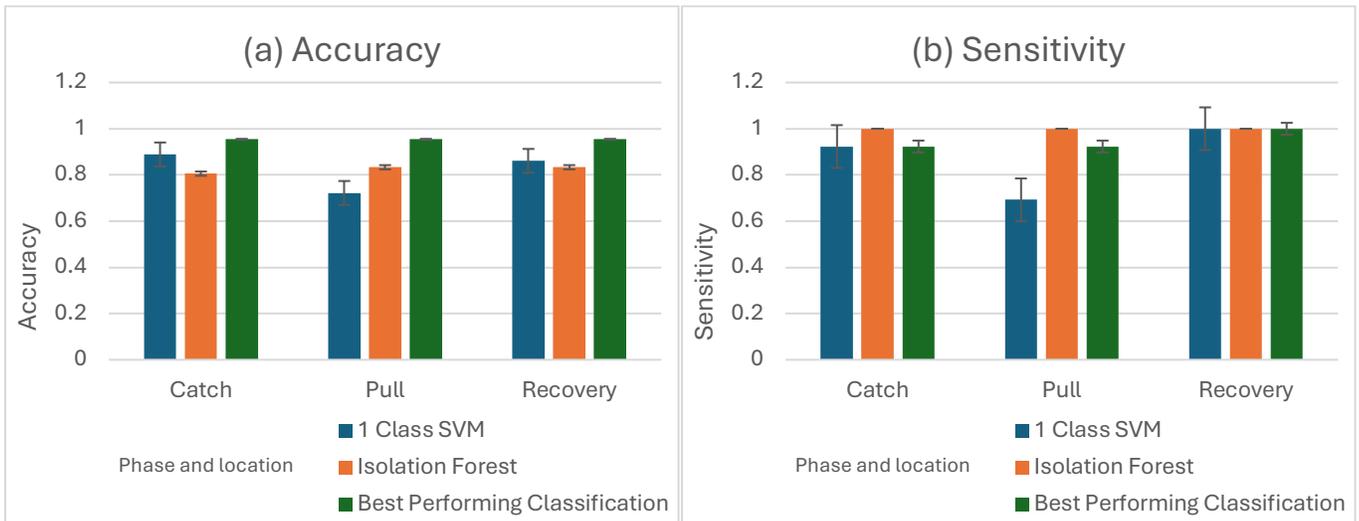

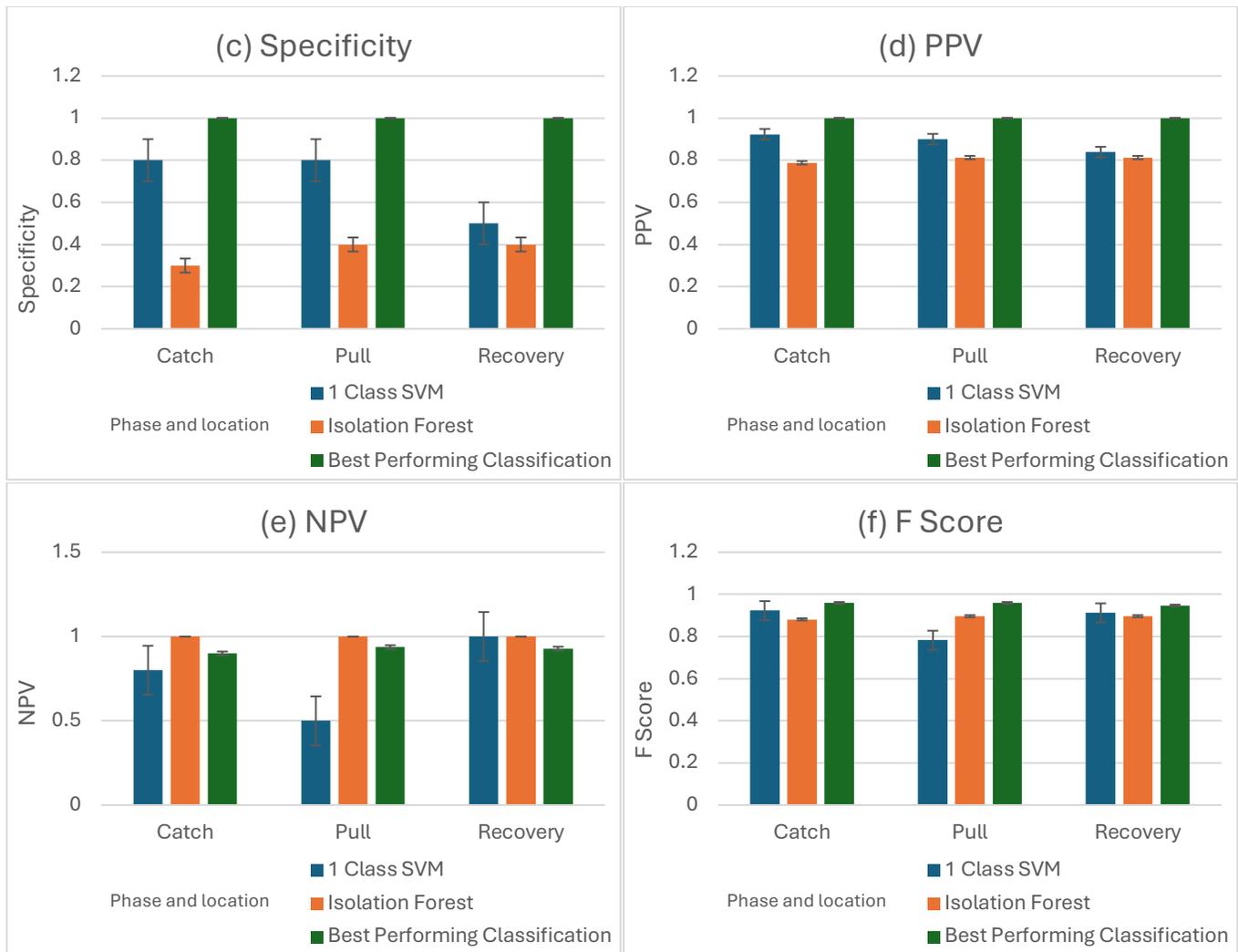

**Figure 15: The performance metrics for anomaly detection models; Sensitivity(a), Specificity(b), PPV (c), NPV (d), Accuracy(e) and F Score (f). The error bars represent standard error values.**

*User-Facing Feedback System*

A user-facing feedback system was developed to enable paddlers to upload data files and interact with the model, as illustrated in **Figure 16**. Users would upload five files: three from the phone and one from each smartwatch. The system then automatically analyzes the data, presenting quantitative results through pie charts and optimal stroke percentages, alongside qualitative feedback generated by an LLM based on the raw graph data.

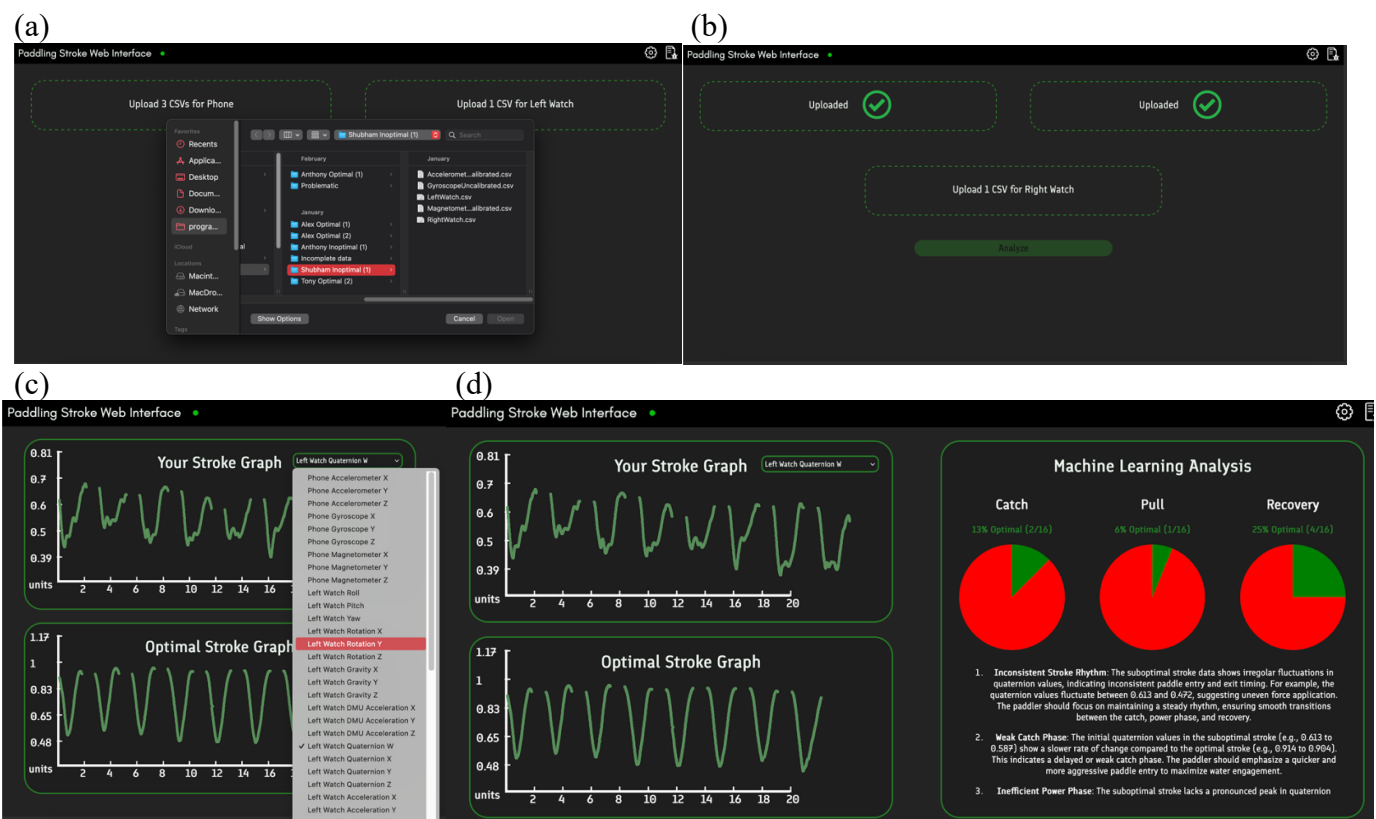

**Figure 16: Screenshots of the user-facing feedback system. File uploading (a) (b), Adjacent graphs of selected features (c), quantitative and qualitative LLM feedback (d).**

**Discussion**

*Principal Results*

This preliminary pilot study suggests that a smartphone and smartwatch-based approach can provide predictive power for paddling stroke quality assessment.

Our analysis focused on three key areas. First, we compared binary classification models using features from all devices, supplemented by anomaly detection model evaluation. Second, we assessed model performance based on data from devices mounted on different body locations (left wrist, right wrist, and bicep) analyzed independently. Finally, we conducted feature importance analysis to identify the most relevant engineered features for stroke quality. These analyses offer insights into optimal device placement and feature selection for future research.

Our analysis evaluated four classification models using summary statistic features. Across all three stroke phases, models achieved high accuracy (~95%) and F-scores between 92% and 96%. Although performance varied slightly among models, the small sample size limits definitive conclusions. Models performed best on the recovery phase, followed by the pull phase, and lastly the catch phase, suggesting that stroke quality may be more affected during paddle removal. However, this also indicates that the thresholding technique for segmentation may have been less effective for the catch phase, despite its recognized importance in stroke technique. Anomaly detection models showed better performance on the catch and recovery phases than the pull phase, phases critical to overall stroke quality, implying they may be more sensitive to stroke errors. Nevertheless, anomaly detection models generally underperformed compared to binary classifiers, suggesting the latter captured broader stroke data patterns. Due to a small sample size, firm conclusions regarding model superiority remain premature.

Analysis of models trained on data from individual sensors revealed advantages of wrist-mounted devices over bicep-mounted ones. The left wrist-mounted device produced the best-performing model, closely followed by the right wrist-mounted device. This likely is an effect of the wrists' proximity to the paddle and their greater range of motion compared to the bicep. These findings suggest that wrist-mounted sensors capture distinctive movement patterns relevant to stroke quality that bicep-mounted sensors may miss. Alternatively, certain data streams unique to smartwatches but absent in smartphones might be linked to higher model performance, although such a proposition was not supported by the feature importance analysis.

The feature importance analysis pinpointed some features as significant, but no specific trend could be observed. This may be attributed to the small sample size of this study.

These findings indicate that motion data from smartphones and smartwatches holds potential for assessing stroke quality using both binary classification and anomaly detection models. While anomaly detection models performed adequately, they did not surpass the binary classifiers. Models trained on bicep-mounted phone data showed reasonable performance but were outperformed by those using wrist-mounted smartwatch data. This disparity likely reflects the greater relevance of wrist positioning or the higher data quality from smartwatches compared to smartphones.

*Comparison to previous works*

This preliminary study advances prior research by introducing a novel approach to paddling stroke quality assessment using data from common smartwatches and smartphones. It offers actionable feedback through quantitative classification metrics and qualitative insights generated by a large language model. Unlike prior methods relying on specialized, costly sensors or high-quality video, this approach democratizes access by leveraging widely available devices and a user-friendly web interface. The results demonstrate that smartphone and smartwatch data can predict stroke quality despite variations in user physique, indicating potential for enhanced accuracy if individual physical differences are incorporated.

Previous studies by Liu et al. (3), McDonnell et al. (4), and Gomes et al. (5) analyzed stroke dynamics using IMU sensors mounted on paddles; however, these relied on specialized equipment rather than commonly available devices. Video-based assessments by Sánchez et al. (6) and Tay et al. (7) successfully captured extensive data but were limited by environmental conditions and required dedicated camera operators to maintain unobstructed views. Moreover, these studies primarily focused on data analysis for researchers, offering limited actionable feedback to users. Our best-performing models demonstrate notable improvements over these approaches, highlighting the effectiveness of the methodological refinements in this study.

While our preliminary pilot study differs significantly from these prior works in terms of device type used, and user-facing-feedback importance, the data collection methods and analysis techniques of previous works strongly influenced this study's design.

*Limitations & Future work*

This study has several limitations that should be considered in related future studies. First, the sample size of 66 strokes from 4 participants was extremely small and is likely insufficient to generalize the findings to paddlers who may use the system without having provided training data. Thus, the system is unlikely to perform well when used by a diverse population. Importantly, the system considered no demographic data from the participants, including gender, age, and nationality. Biometric and physical data was also not considered, as height and weight may have influenced observed patterns. Critically, handedness was not considered, as all participants were asked to complete right-side paddling regardless of their dominant hand. In addition, the close succession of trials (within 15 minutes of each other) may have led to the second trial being influenced by the paddler's soreness or tiredness. Additionally, the thresholding-based stroke segmentation process was questionable in reliability and led to some strokes being discarded when the process failed. Future research may focus on following a similar approach using everyday devices but performing more robust analysis with a larger stroke sample size of participants and strokes. In addition, it should employ a larger number of sensors and consider positions closer to the paddle. Real-time feedback should also be a key point of development to provide more comprehensive and actionable paddling feedback.

## Contributors

Conceptualization: SP, AL; Data collection: AL, SP, PB; Web development: SP, AH, AL, PB; Writing: SP; Data analysis: SP, AL, PB; All authors had full access to the data used in the study and had final responsibility for the decision to submit for publication.

## Declarations of Interests

We declare no competing interests.

## Data Sharing

An anonymized version of the data used in this study and the models trained will be released upon publication.